\documentclass[times,twocolumn,final]{elsarticle}
\usepackage[
  left=2.35cm,
  right=2.35cm,
  top=1.8cm,
  bottom=2.0cm,
  columnsep=0.6cm
]{geometry}

\usepackage{framed,multirow}

\usepackage{amssymb}
\usepackage{latexsym}

\usepackage{url}
\usepackage{xcolor}

\usepackage{pifont}
\newcommand{\cmark}{\ding{51}}%
\newcommand{\xmark}{\ding{55}}%

\usepackage[hidelinks]{hyperref}
\definecolor{newcolor}{rgb}{.8,.349,.1}

\usepackage[ruled,vlined,algo2e]{algorithm2e}
\SetKwComment{Comment}{$\triangleright$\ }{}

\usepackage{mathtools}
\usepackage{amsmath}

\newcommand\Tstrut{\rule{0pt}{2.0ex}}         
\newcommand\Bstrut{\rule[-0.9ex]{0pt}{0pt}}   

\begin{document}


\begin{frontmatter}

\title{PPDM: Pixel Puzzling Diffusion Model for Speed and Memory Efficient Volumetric Medical Image Translation}

\author[1,2]{Tianqi Chen} 
\author[2]{Jun Hou}
\author[2]{Yinchi Zhou}
\author[2,3]{James S. Duncan}
\author[2,3]{Chi Liu}
\author[1]{Bo Zhou\corref{cor1}} \ead{bo.zhou@northwestern.edu}
\cortext[cor1]{Corresponding author.}

\address[1]{Department of Radiology, Northwestern University, Chicago, IL, USA}
\address[2]{Department of Biomedical Engineering, Yale University, New Haven, CT, USA}
\address[3]{Department of Radiology and Biomedical Imaging, Yale School of Medicine, New Haven, CT, USA}




\begin{abstract}
Diffusion models have demonstrated superior fidelity for medical image–to–image translation, but their extension to high-resolution 3D volumes is severely constrained by prohibitive computational cost and GPU memory requirements. Existing memory-efficient strategies often compromise global volumetric consistency or fine anatomical detail. In this work, we propose the Pixel Puzzling Diffusion Model (PPDM), a simple and effective framework for memory- and speed-efficient 3D medical image translation. PPDM introduces a reversible pixel puzzle–unpuzzle operator that trades spatial resolution for channel dimensionality, substantially reducing activation memory while preserving global context. To further improve efficiency and stability, we adopt a direct bridge diffusion formulation that starts from the conditional input rather than pure noise, enabling the model to focus on task-relevant residuals. In addition, a puzzle-gradient loss is incorporated to enforce spatial coherence and suppress grid-like artifacts introduced by spatial rearrangement. We evaluate PPDM on multiple challenging 3D medical image translation tasks, including low-count PET denoising, joint PET denoising and attenuation correction, and cross-modal MRI translation. Across all tasks, PPDM consistently matches or outperforms full 3D diffusion models while reducing training GPU memory usage by up to an order of magnitude and significantly accelerating inference, and it outperforms existing memory-efficient diffusion approaches based on latent compression or frequency decomposition. These results demonstrate that PPDM provides a practical and scalable solution for high-fidelity 3D diffusion-based medical image translation under limited computational resources.

\end{abstract}


\end{frontmatter}


\section{Introduction}
Various medical image–to–image (I2I) translation tasks play a central role in modern medical imaging pipelines by enabling integration, enhancement, and substitution across modalities such as positron emission tomography (PET), computed tomography (CT), and magnetic resonance imaging (MRI). In PET imaging, low-dose PET can be translated to corresponding full-dose PET to reduce radiation exposure while preserving quantitative accuracy \citep{xue2025deep}. In CT, few-view acquisitions can be mapped to full-view scans to enable faster imaging and a lower radiation dose \citep{zhou2021limited, zhang2018sparse}. In MRI, cross-contrast image translation (e.g., T1 $\rightarrow$ T2 or T2 $\rightarrow$ FLAIR) allows comprehensive diagnostic assessment without extending scan protocols \citep{yang2020mri, welander2018generative}, or accelerate image acquisition time \citep{zhou2020dudornet}. Beyond intra-modality translation, cross-modality I2I tasks can supplement missing contextual information without introducing additional scanning procedures. For example, synthetic CT generation from PET enables CT-free attenuation correction and reconstruction in PET imaging \citep{zhou2024pour, chen20252, xie2024dose}. Similarly, CT images synthesized from MRI can provide complementary information \citep{nie2018medical, wolterink2017deep}. This is particularly important in radiation therapy, where both CT and MRI are routinely required: CT provides electron density information essential for dose calculation \citep{shafai2019dose, palmer2021synthetic}, while MRI offers superior soft-tissue contrast for accurate delineation of tumors and surrounding organs at risk \citep{aouadi2025towards, vellini2025deep}. 

With the advent of deep learning, modern machine learning approaches have achieved remarkable progress in I2I translation and image generation, demonstrating strong performance in both natural and biomedical imaging domains. Convolutional neural network (CNN)–based models, such as U-Net \citep{ronneberger2015u} and conditional generative adversarial networks (cGANs) \citep{isola2017image}, have been widely applied to medical I2I tasks by learning direct conditional mappings between input and target images. In PET imaging, PT-WGAN \citep{gong2020parameter} denoises low-dose PET by transferring generator parameters pretrained on standard-dose data, while CNN-based approaches \citep{hu2020simultaneous} perform attenuation correction, scatter correction, and denoising without requiring CT. In MRI, CNN-based methods enable cross-field-strength translation from 3T to 7T \citep{nie2018medical} and cross-contrast synthesis to reduce scan time while preserving anatomical structure and diagnostic fidelity \citep{yang2020mri}. These approaches have also been extended to cross-modality translation, including PET-to-CT \citep{li2022tcgan}, PET-to-MRI \citep{bazangani2022fdg}, and MRI-to-CT synthesis \citep{xing2024cross}. Despite their success, CNN-based methods often suffer from oversmoothing, limited receptive fields, and insufficient modeling of long-range spatial dependencies. These limitations can lead to degraded boundary sharpness and compromised global anatomical consistency, particularly in volumetric medical images.

Diffusion models \citep{ho2020denoising} have recently emerged as a compelling alternative for conditional image synthesis and I2I translation. By modeling the data distribution through a stochastic denoising process, diffusion models exhibit more stable training dynamics and superior image fidelity compared with traditional CNN-based approaches. For example, to mitigate the randomness induced by noise-initialized diffusion in I2I tasks, recent works reformulate diffusion processes to start directly from the input image. \citep{li2023bbdm} introduces Brownian Bridge Diffusion, which directly translates between source and target domains, while \citep{liu20232} employs Schrodinger Bridge theory to achieve optimal distributional transport. These formulations substantially reduce uncertainty and have shown strong performance in medical I2I translation. Recent studies have demonstrated their effectiveness in medical imaging applications such as PET denoising \citep{gong2024pet, shen2023pet, pan2024full}, CT-free attenuation correction and reconstruction \citep{chen20252}, and MRI modality translation \citep{xing2024cross}. Beyond same-modality translation, diffusion models have also been successfully applied to cross-modality image synthesis. For example, \citet{lyu2022conversion} translate CT to MRI using a diffusion and score-based framework, \citet{graf2023denoising} generate synthetic CT from MRI to enable automated spine segmentation, and \citet{choo2024slice} employ a Brownian bridge diffusion model to produce slice-consistent 3D MRI volumes from CT. Across these tasks, diffusion-based approaches consistently yield sharper structural details and more anatomically faithful results than conventional methods.

However, most successful deep-learning-based I2I methods are developed for 2D image data and do not directly extend to high-resolution 3D medical volumes. In particular, diffusion models have an extremely high computational cost due to their iterative sampling process, which requires multiple time steps. While this cost is manageable for 2D images, it becomes prohibitive for volumetric data, where both computation and activation memory scale approximately cubically with spatial resolution ($\mathcal{O}(N^3)$), quickly exceeding the capacity of commodity GPUs.

Recent work has explored strategies to mitigate the memory and computational burden of 3D diffusion models. Existing approaches primarily fall into two categories. The first reduces the dimensionality of the network input. For example, partition-and-aggregate methods generate only part of the volume at a time, either using slice-wise (2D or 2.5D) processing or tiled 3D patches, and subsequently stitch predictions into a full 3D image. 
Other approaches reduce the dimensionality of the original high-resolution volume itself. Huang et al. \citep{huang2024wavedm} proposed a wavelet-based diffusion model (WDM) that decomposes 3D images into multiple low-resolution sub-band channels. An alternative strategy compresses the high-dimensional input into a lower-dimensional latent space using an encoder, as in latent diffusion models such as ALDM \citep{kim2024adaptive}. In these methods, diffusion is performed in the latent space, and a decoder reconstructs the denoised features back to the original image domain.

However, these methods have notable limitations. Partition-and-aggregate approaches, while leveraging mature 2D backbones and fitting within limited GPU memory, often fail to capture long-range anatomical context, introduce boundary artifacts, and degrade global structural consistency. Wavelet-based approaches decompose images into frequency components, concentrating fine anatomical structures in high-frequency bands that diffusion models struggle to reconstruct accurately. As a result, they tend to produce overly smooth textures, blurred details, and diminished anatomical boundaries. Moreover, the fixed number of wavelet sub-bands limits flexibility across different imaging resolutions and modalities. Latent diffusion methods, although memory-efficient, operate in a compressed latent space that substantially reduces effective spatial resolution and relies heavily on a robust decoder to recover high-fidelity anatomy. Such aggressive compression can obscure clinically important details, and training reliable 3D encoders remains challenging due to limited dataset sizes and substantial anatomical variability.

A second category of approaches focuses on architectural simplification by designing lightweight 3D networks with reduced parameter counts. While this alleviates memory constraints, reduced model capacity often limits reconstruction fidelity. Moreover, as spatial resolution increases, computation and activation memory still scale cubically, making high-resolution volumetric diffusion modeling inherently difficult. 

To overcome these challenges, we propose a novel, simple, and efficient framework that accelerates both training and inference for 3D medical image translation while substantially reducing GPU memory consumption, without sacrificing global volumetric consistency. Specifically, we introduce the Pixel Puzzling Diffusion Model (PPDM), a memory-efficient diffusion framework that synergistically integrates three key components. First, PPDM employs a pixel-wise puzzle–unpuzzle mechanism that trades spatial resolution for channel dimensionality, dramatically reducing activation memory and enabling a larger effective receptive field on commodity GPUs. Second, conventional DDPMs perform image translation by denoising from pure Gaussian noise, introducing a large distributional gap between the initial state and the target image and requiring many denoising steps. In contrast, our direct diffusion formulation starts from the input image itself, where the gap between input and output is substantially smaller, allowing the model to focus only on task-relevant corrections. Finally, to constrain boundaries of anatomical organs during diffusion in the puzzle space and mitigate channel-wise inconsistencies introduced by spatial rearrangement, we proposed a puzzle-gradient loss that constrains the anatomical boundaries and enhances volumetric translation accuracy. Extensive experiments on multiple 3D medical image translation tasks demonstrate that PPDM achieves state-of-the-art performance while requiring significantly lower computational and memory resources.

\begin{figure*}[htb!]
\centering
\includegraphics[width=0.76\textwidth]{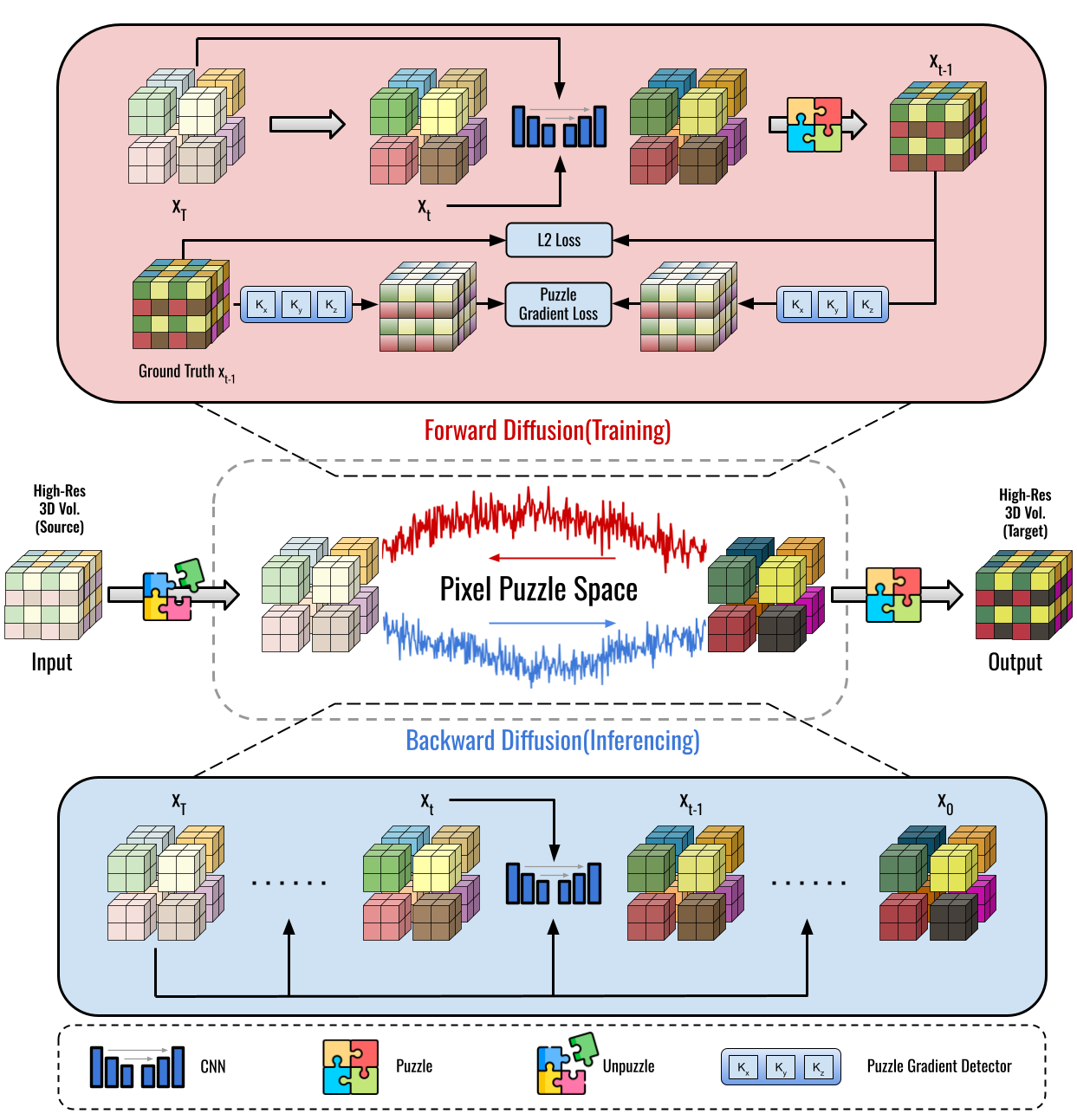}
\caption{Overview of the proposed Pixel Puzzle Diffusion Model (PPDM).
The input 3D volume is first transformed using the pixel unpuzzle operator to trade spatial resolution for channel dimensionality, enabling memory-efficient diffusion processing. A direct bridge diffusion model predicts residuals conditioned on the input, while a puzzle-gradient loss enforces spatial coherence and suppresses grid-like artifacts in the forward process. The final output is reconstructed by applying the puzzle operator to restore full spatial resolution.}
\label{fig:pipeline}
\end{figure*}
\section{Methods}
The overall pipeline of the Pixel Puzzle Diffusion Model (PPDM) is illustrated in Figure~\ref{fig:pipeline}. This section proceeds as follows: Section~2.1 introduces the 3D pixel–puzzle operator for memory-efficient volumetric processing; Section~2.2 formulates the PPDM training and inference procedures, including the bridge diffusion and puzzle gradient losses; Section~2.3 details the data preparation; Section~2.4 describes the implementation details; and Section~2.5 outlines the evaluation metrics and baselines.

\subsection{3D Pixel Puzzle Operator}

The proposed 3D pixel puzzle operator consists of two complementary transformations: the \textit{pixel unpuzzle} operator and the \textit{pixel puzzle} operator. These operators rearrange 3D voxel grids to exchange spatial resolution for channel dimensionality, enabling memory-efficient processing in 3D diffusion models.

\noindent\textbf{Pixel Unpuzzle:}
Given an input matrix $X \in \mathbb{R}^{C \times D \times H \times W}$, the pixel unpuzzle operator rearranges voxels into a lower spatial resolution matrix
\[
X_L \in \mathbb{R}^{(C r^3) \times \frac{D}{r} \times \frac{H}{r} \times \frac{W}{r}},
\]
where $r$ is the spatial down-scaling factor. The mapping is defined as
\begin{equation}
    X_L[c', x', y', z'] = X[c, x, y, z],
\end{equation}
with the index relationships:
\begin{align}
    c' &= c \cdot r^3 + (x \bmod r) + (y \bmod r)\, r + (z \bmod r)\, r^2, \\
    x' &= \left\lfloor \frac{x}{r} \right\rfloor,\qquad
    y' = \left\lfloor \frac{y}{r} \right\rfloor,\qquad
    z' = \left\lfloor \frac{z}{r} \right\rfloor.
\end{align}
where $c, x, y, z \in \mathbb{Z}$ with
$c \in [0, C)$, $x \in [0, D)$, $y \in [0, H)$, and $z \in [0, W)$.

We denote the pixel unpuzzle operation as
\[
X_L = \mathcal{P}^{-1}(X, r),
\]
where $\mathcal{P}^{-1}$ represents the 3D pixel unpuzzle transformation.

\vspace{0.1cm}
\noindent\textbf{Pixel Puzzle:}
The reverse transformation restores the original spatial structure:
\[
X = \mathcal{P}(X_L, r),
\]
where $\mathcal{P}$ is the inverse pixel puzzle operator that maps the matrix $x_L$ back to the original shape $C \times D \times H \times W$.


\subsection{Puzzled-space Diffusion Model}
We adopt a direct diffusion process for our PPDM. Instead of traditional DDPM \citep{ho2020denoising} starting from pure random Gaussian noise, the conditional input $y$ is fixed as the starting point at time step T, and the forward process constructs a stochastic bridge between the target $x_0$ and $y$. Specifically, at each timestep $t \in \{0,1,\dots,T\}$, the forward process is defined as:

\begin{equation}
x_t = (1 - m_t) x_0 + m_t y + \sqrt{\delta_t} \epsilon, \quad \epsilon \sim \mathcal{N}(0, \mathbf{I}),
\end{equation}

where $m_t = \tfrac{t}{T}$ is the interpolation weight and $\delta_t = 2(m_t - m_t^2)$ controls the noise variance. As $t$ increases, $x_t$ transitions smoothly from $x_0$ to $y$, forming a stochastic bridge.

\noindent\textbf{Training.}
The denoising model $f_\theta(y, x_t, t)$ is trained to predict the stochastic residual term $\mu_t$ defined as:

\begin{equation}
\mu_t = m_t (y - x_0) +  \sqrt{\delta_t} \epsilon.
\end{equation}

The training objective is to minimize the discrepancy between the predicted and true residuals. A pixel-wise L2 loss is employed:

\begin{equation}
\mathcal{L}_{p} =  \left\| f_\theta(y, x_t, t) - \mu_t \right\|^2.
\end{equation}

While the pixel puzzle operator enables substantial memory savings by distributing spatial information across channels, it may introduce inconsistencies between channel predictions, potentially leading to subtle grid-like artifacts in the reconstructed image. To address this, we additionally introduce a puzzle-gradient loss that promotes spatial coherence across puzzled voxel blocks and suppresses such artifacts.

Given a 3D input tensor $X \in \mathbb{R}^{1 \times D \times H \times W}$, 
we compute spatial gradients using $s$ 3D Sobel kernels in different directions:
\begin{equation}
K = \{K_1,\, K_2,\, \dots,\, K_s\},
\end{equation}
where each kernel $K_i \in \mathbb{R}^{1 \times n \times n \times n}$ encodes a specific 3D spatial orientation, with $n$ denoting the kernel size.

Each kernel is applied to the input volume using 3D convolution in its corresponding direction. 
The final edge magnitude is computed as the $L_2$ norm of all gradient responses:
\begin{equation}
E(X) = \sqrt{\sum_{i=1}^{s} (K_i * X)^2 }.
\end{equation}

Therefore, the puzzle gradient loss is computed as:

\begin{equation}
\mathcal{L}_{\text{g}} =  \left\| E(\mathcal{P}(f_\theta(y, x_t, t), r)) -  E(\mathcal{P}(\mu_t, r)) \right\|^2,
\end{equation}

The total training loss is:

\begin{equation}
\mathcal{L}_{total} = \mathcal{L}_{p} + \lambda \mathcal{L}_{\text{g}}.
\end{equation}

\noindent\textbf{Inference:} 
At inference time, once the model $f_\theta(\cdot)$ has been trained, we perform sampling by backward diffusion process. Given the conditional input volume $y$, we first apply the 3D pixel unpuzzled operator to obtain its low spatial resolution representation
\begin{equation}
y_L = \mathcal{P}^{-1}(y, r),
\end{equation}
which serves as the starting point for memory-efficient diffusion. The reverse process is initialized as $x_T = y_L$, and the model iteratively predicts the residual $\mu_t$ and reconstructs $x_{t-1}$ from $x_t$ until reaching the final estimate $x_0$. The reverse update rule is
\begin{equation}
x_{t-1} = c_{x,t} \, x_t + c_{y,t} \, y_L - c_{\epsilon,t} \, f_\theta(y_L, x_t, t),
\end{equation}
where
\begin{equation}
c_{x,t} = \frac{\delta_{t-1}}{\delta_t} \cdot \frac{1-m_t}{1-m_{t-1}}
       + \frac{\delta_{t \mid t-1}}{\delta_t} \cdot (1-m_{t-1}),
\end{equation}
\begin{equation}
c_{y,t} = m_{t-1} - m_t \cdot \frac{1-m_t}{1-m_{t-1}} \cdot \frac{\delta_{t-1}}{\delta_t},
\end{equation}
\begin{equation}
c_{\epsilon,t} = (1-m_{t-1}) \cdot \frac{\delta_{t \mid t-1}}{\delta_t},
\end{equation}
and
\begin{equation}
\delta_{t \mid t-1} = \delta_t - \delta_{t-1} \frac{(1 - m_t)^2}{(1 - m_{t-1})^2}.
\end{equation}

Once the diffusion process reaches $x_0$, the prediction still resides in the low spatial resolution domain. Therefore, we apply the pixel puzzle operator to restore the original volumetric resolution:
\begin{equation}
\hat{x} = \mathcal{P}(x_0, r),
\end{equation}
yielding the final high-resolution reconstructed output.

\subsection{Dataset Preparation}
We train and evaluate our method on three different datasets. The details are described as follows:

\noindent\textbf{Shanghai Ruijin Hospital Dataset \citep{xue2022cross}:} This dataset is for the low-count PET denoising task, which includes a total of 279 subjects, with 253 used for training and 26 for testing. All scans were acquired using a United Imaging Healthcare uExplorer total-body PET/CT system. Image reconstruction was performed using the OSEM algorithm with 4 iterations and 20 subsets. For each subject, we manually cropped the head region as the central region of interest, resulting in a voxel size of $128 \times 128 \times 64$. Low-dose PET data were generated by downsampling the list-mode data to a $1\%$ low-count level through list-mode rebinning.

\noindent\textbf{Yale New Haven Hospital (YNHH) Dataset:} This dataset is for a simultaneous low-count PET attenuation correction and denoising task, which includes a total of 98 subjects, with 65 used for training and 33 for testing. All scans were acquired using a Siemens Biograph mCT PET/CT scanner at Yale New Haven Hospital. Low-count PET data were generated by uniformly downsampling the list-mode data to $5\%$ count level. The non-attenuation-corrected low-count PET images (NAC-LDPET) were reconstructed from the downsampled list-mode data using the ordered-subsets expectation maximization (OSEM) algorithm with 2 iterations and 21 subsets, without attenuation correction. The attenuation-corrected full-count PET images (AC-SDPET) were reconstructed using the original list-mode data with the same OSEM settings but with attenuation correction applied. For each subject, we manually cropped the head region as the central region of interest, resulting in a voxel size of $96 \times 96 \times 96$.

\noindent\textbf{IXI Dataset \citep{IXIDataset2018}:} This dataset is for a cross-modal MRI translation task, which includes a total of 318 subjects, with 295 used for training and 23 for testing. The IXI dataset was collected at Guy’s Hospital, London, using a Philips 1.5T MR scanner. It consists of paired proton density (PD) and T2-weighted MRI scans. We used the T2-weighted images as the input and the PD-weighted images as the target output for model training. All MRI volumes were resampled and cropped to a uniform voxel size of $128 \times 128 \times 64$.

\subsection{Implementation Details}
All experiments were conducted on an NVIDIA H100 GPU. Models were trained with a batch size of 8 using the Adam optimizer and a learning rate of $1\times10^{-5}$ until convergence. An exponential moving average (EMA) with a decay rate of 0.9999 was applied to stabilize training. Both training and inference employed a diffusion process with 1000 timesteps. By default, the proposed PPDM was trained and evaluated using $\lambda = 0.1$, kernel size $n = 3$, and spatial down-scale factor $r = 2$.

\subsection{Evaluation Metrics and Baselines}
The quality of the translated images was evaluated at the image level using three image-quality metrics, including Peak Signal-to-Noise Ratio (PSNR), Structural Similarity Index (SSIM), and Normalized Mean Square Error (NMSE). We compare our method to several state-of-the-art baselines that specifically proposed to address the memory challenges in the diffusion model, including 3D BBDM \citep{li2023bbdm}, PatchDDM \citep{bieder2024memory}, WDM \citep{huang2024wavedm}, and ALDM \citep{kim2024adaptive}.

\begin{figure*}[htb!]
\centering
\includegraphics[width=0.92\textwidth]{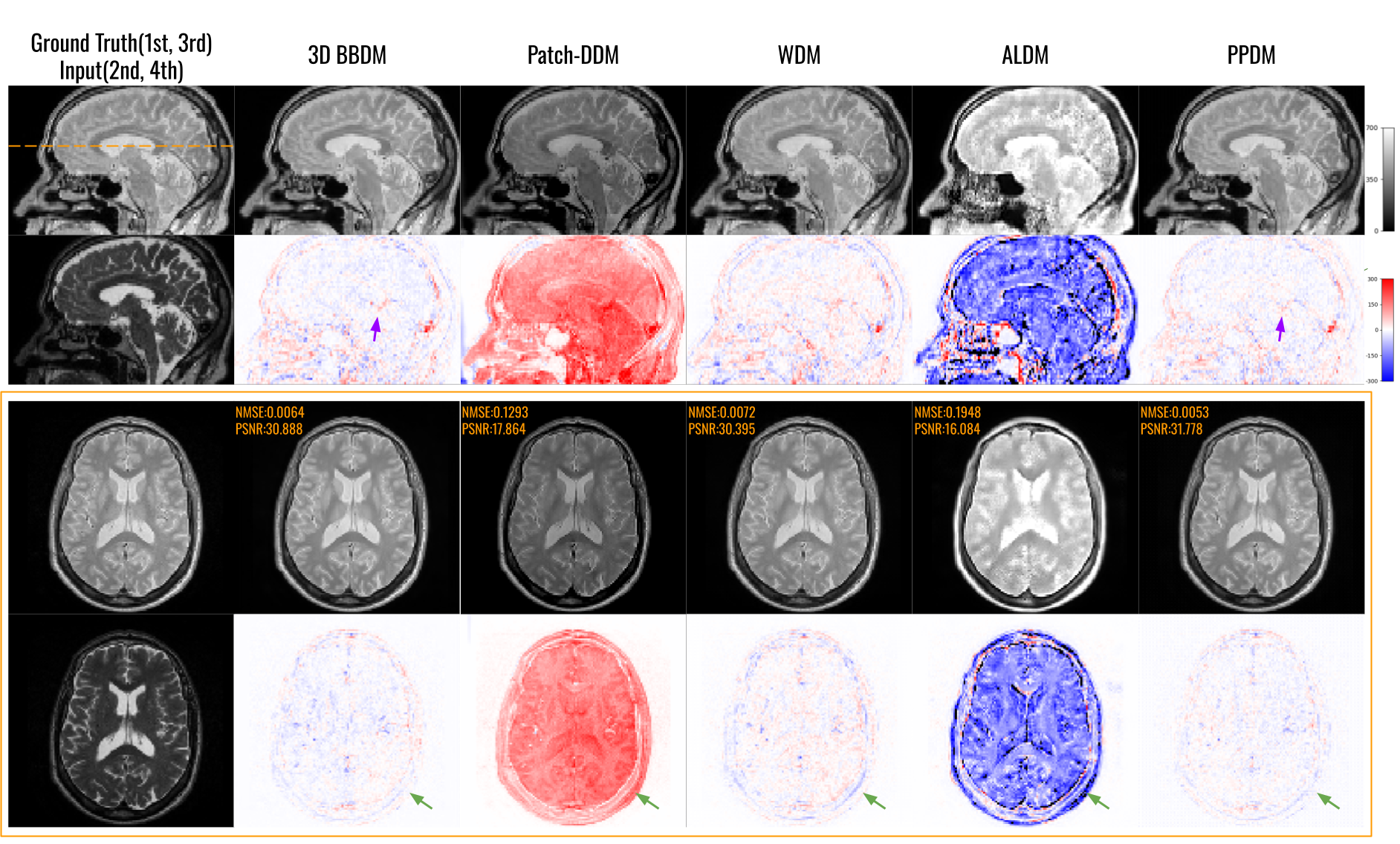}
\caption{Visual comparison of PD-weighted MRI generated from T2-weighted MRI using different methods. Coronal slices (first row) and their corresponding error maps (second row), as well as axial slices (third row) and their corresponding error maps (fourth row), are shown. NMSE and PSNR are computed at the full 3D volume level and reported at the top of the axial slices.}
\label{fig:mri}
\end{figure*}
\section{Results}

\begin{figure*}[htb!]
\centering
\includegraphics[width=0.9\textwidth]{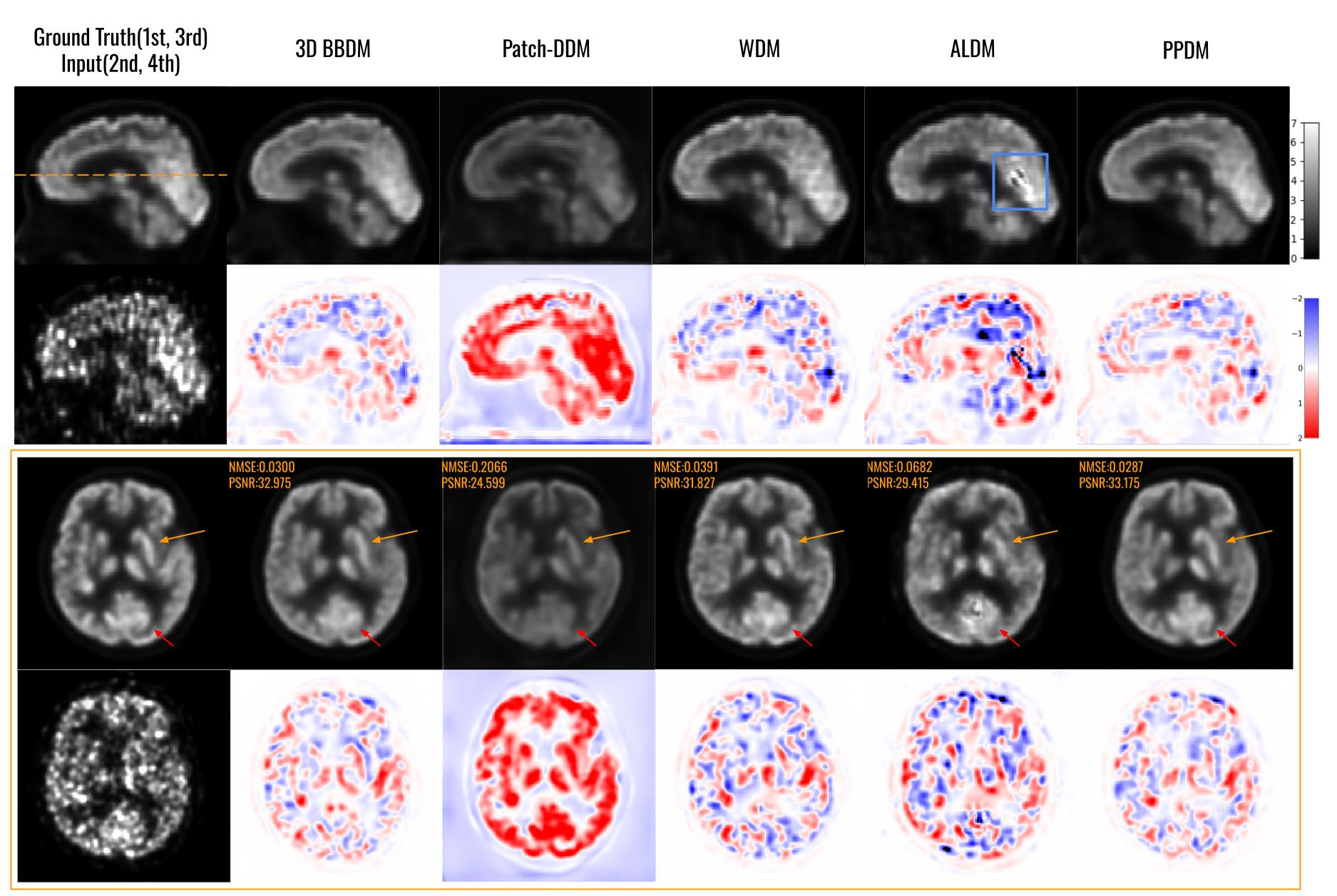}
\caption{Visual comparison of standard-dose PET generation from different methods under 1\% low-dose settings. The coronal view results (1st row) and their error maps (2nd row), and axial view results (3rd row) and their error maps(4th row) are shown. NMSE and PSNR values are calculated at the 3D volume level and shown at the bottom of axial slices.}
\label{fig:pet_denoise}
\end{figure*}


\begin{figure}[htb!]
\centering
\includegraphics[width=0.50\textwidth]{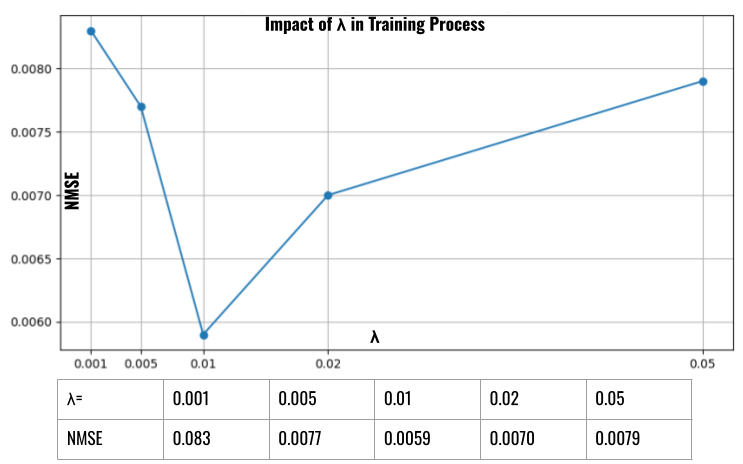}
\caption{Ablation study of the puzzle-gradient loss weight $\lambda$ on the IXI dataset. NMSE is plotted as a function of $\lambda$, with the corresponding numerical values summarized in the accompanying table. The best performance is achieved at $\lambda = 0.01$.}
\label{fig:abl_plot_lam}
\end{figure}

\begin{table*} [htb!]
\tiny
\centering
\caption{Quantitative evaluation on Yale PET head (5\% low-count) denoising and attenuation correction, IXI MRI translation, and Shanghai PET head 3D (1\% low-count) denoising tasks. We compare 3D BBDM, PatchDDM, WDM, ALDM, and the proposed PPDM using NMSE, SSIM, PSNR, training GPU memory consumption, and inference TFLOPs. Best results are highlighted in \textbf{bold}.}

\label{tab:compare_method}
\resizebox{0.9\textwidth}{!}{
    \begin{tabular}{l|c|c|c|c|c}
        \hline
        \multirow{2}{*}{\textbf{Evaluation}} &\multicolumn{4}{c}{\textbf{PET AC + Denoising}}
        \Tstrut\Bstrut\\
        \cline{2-6}
        & NMSE & SSIM & PSNR &Training Memory(GB)  & TFLOPs\Tstrut\Bstrut\\
        \hline
        3D BBDM              & $0.0131 \pm  0.0043$ & $0.9852 \pm 0.0047$ & $32.774 \pm 2.501$ & $49.07$ & $18160$ \\
        PatchDDM            & $0.2518 \pm 0.0225$ & $0.8295 \pm 0.0279$ & $19.732 \pm 2.624$ & $1.22$ &$1054$\Bstrut\\
        WDM                       & $0.0206 \pm 0.0113$ & $0.9810 \pm 0.0053$ & $31.092 \pm 2.878$ & $6.94$ &$2280$\\
        ALDM                 & $0.0606 \pm 0.0381$ & $0.9335 \pm 0.0494$ & $26.682 \pm 3.473 $ & $1.36$ & $163$\\
        PPDM      & $\mathbf{0.0130 \pm 0.0040}$ & $\mathbf{0.9854 \pm 0.0043}$ & $\mathbf{32.774 \pm 2.515}$ & $6.94$ &$2280$\Tstrut\\
        \hline

        \hline
        &\multicolumn{4}{c}{\textbf{IXI T2 to PD}}
        \Tstrut\Bstrut\\
        \hline
        3D BBDM         & $0.0070 \pm 0.0014$ & $0.9938 \pm 0.0009$ & $31.720 \pm 1.975$ &  $59.68$ &$21610$\Tstrut\\
        PatchDDM                 & $0.1182 \pm 0.0074$ & $0.9121 \pm 0.0092$ & $19.368 \pm 2.509$ & $1.05$  &$890$ \Bstrut\\
        WDM                & $0.0080 \pm 0.0015$ & $0.9919 \pm 0.0011$ & $31.109 \pm 2.013$ & $8.06$ & $2700$\\
        ALDM                 & $0.3241 \pm 0.2298$ & $0.8788 \pm 0.0592$ & $15.580 \pm 1.459$ & $1.36$ &$163$\\
        PPDM    & $\mathbf{0.0059 \pm 0.0013}$ & $\mathbf{0.9934 \pm 0.0010}$ & $\mathbf{32.436 \pm 1.899}$ & $16.90$ &$5460$
        \Bstrut\\
        \hline

        \hline
        &\multicolumn{4}{c}{\textbf{{PET Denoising}}}
        \Tstrut\Bstrut\\
        \hline
        3D BBDM & $0.0229 \pm 0.0043$ & $0.9876 \pm 0.0029$ & $30.762 \pm 1.617$ &  $59.68$ &$21610$\Tstrut\\
        PatchDDM         & $0.3296 \pm 0.0677$ & $0.7752 \pm 0.0451$ & $19.211 \pm 2.691$ & $1.05$  &$890$\\
        WDM          & $0.0304 \pm 0.0055$ & $0.9841 \pm 0.0035$ & $29.534 \pm 1.658$ & $8.06$ & $2700$\\
        ALDM         & $0.0566 \pm 0.0267$ & $0.9679 \pm 0.0171$ & $27.052 \pm 2.309$ & $1.36$ &$163$\\
        PPDM & $\mathbf{0.0223 \pm 0.0040}$ & $\mathbf{0.9880 \pm 0.0027}$ & $\mathbf{30.868 \pm 1.655}$  & $16.90$ &$5460$\Bstrut\\
        \hline
    \end{tabular}
}
\end{table*}

\subsection{Experimental Results}
In Fig. \ref{fig:mri}, we present qualitative comparison results for the MRI modality translation task, in which T2-weighted images are translated into PD-weighted contrast using different approaches. Among the compared approaches, ALDM and PatchDDM produce the lowest image quality, with relatively low PSNR values of 16.084 dB and 17.864 dB, and high NMSE values of 0.1948 and 0.1293, respectively. Their reconstructions are substantially inaccurate and show clear intensity deviations from the ground truth, including both over- and under-estimated regions. While WDM generates images of relatively higher quality, it still introduces bias in the meninges region (green arrow) 3D BBDM, and our proposed PPDM. Both the 3D BBDM and our PPDM produce the most accurate reconstructions, closely matching the anatomical structures and contrast of the ground truth. Notably, PPDM further improves reconstruction quality over the 3D BBDM, reducing the NMSE from 0.0064 to 0.0053 and increasing the PSNR from 30.888 dB to 31.778 dB. In addition to its superior quantitative performance, PPDM yields the lowest reconstruction error in the quadrigeminal cistern region (purple arrow), as indicated by the error map, demonstrating its enhanced ability to preserve fine anatomical structures.

In Fig. \ref{fig:pet_denoise}, we present a qualitative comparison for the PET denoising task. The 1\% low-dose PET input suffers from severe noise due to extreme count reduction. Similar to MRI modality translation task, PatchDDM and ALDM produce images with the poorest visual quality. In particular, ALDM exhibits pronounced structural distortions in the putamen region (blue box), and PatchDDM result a underestimation of the brain region. WDM also presents structural integrity degradation, as indicated by its relatively higher NMSE of 0.0391. In contrast, both the 3D BBDM and our proposed PPDM reliably recover brain structures and produce results that most closely resemble the ground truth. Notably, PPDM achieves the best quantitative performance in this example, with the lowest NMSE and highest PSNR among all methods. Moreover, PPDM provides superior visualization of the putamen and preserves sharp anatomical boundaries in the cerebellum (red arrows) and cortical (orange arrows) regions, yielding the most faithful reconstruction to the ground truth. 

The quantitative comparisons in Table \ref{tab:compare_method} corroborate the trends observed in the visual results. For the PET denoising and attenuation correction (AC) task, 3D BBDM achieves strong performance with an NMSE of 0.0131, but at the cost of substantially increased training memory consumption. PatchDDM reduces the memory requirement to 1.22 GB; however, this memory efficiency comes with a considerable degradation in reconstruction accuracy, resulting in an NMSE of 0.2518. WDM also offers significant memory savings by reducing the training memory to about 6.94 GB, but this also comes with a decrease in accuracy, yielding an NMSE of 0.0206. Similarly, the latent diffusion model, ALDM, produces relatively large reconstruction errors, with an average NMSE of 0.0606 and SSIM of 0.9335. Notably, our proposed PPDM achieves the best quantitative performance across all evaluated metrics. Specifically, PPDM reduces the NMSE from 0.0131 to 0.0130 compared with the second-best method, 3D BBDM, while simultaneously reducing memory usage by more than 7 times.

Similar results are observed in the IXI MRI translation task. 3D BBDM achieves strong reconstruction performance with an NMSE of 0.0070, but this comes at the cost of substantial training GPU memory consumption, reaching 59.68 GB. Although ALDM and PatchDDM reduce computational cost, they exhibit the poorest performance, with markedly high reconstruction errors corresponding to NMSE values of 0.3241 and 0.1182, respectively. These results indicate unstable and inaccurate modality translation. WDM also substantially reduces training memory consumption to 8.06 GB, but it achieves lower quantitative accuracy than 3D BBDM. In contrast, the proposed PPDM achieves the best overall accuracy, with an NMSE of 0.0059, while requiring substantially lower training memory than 3D BBDM, at 16.90 GB. These results demonstrate that PPDM provides a superior balance between reconstruction fidelity and computational efficiency.

A similar trend is observed for the PET denoising task. Although PatchDDM, ALDM, and WDM reduce training memory usage, they exhibit weaker reconstruction performance than 3D BBDM and PPDM. PatchDDM and ALDM achieve the lowest memory consumption, requiring only 1.05 GB and 1.36 GB, respectively, but produce the weakest results, with high NMSE values of 0.3296 and 0.0566. WDM improves the NMSE to 0.0304 while reducing training memory consumption to 8.06 GB, but it still underperforms 3D BBDM. In contrast, 3D BBDM achieves strong reconstruction accuracy with an NMSE of 0.0229, but requires substantially higher training memory consumption, reaching 59.68 GB. The proposed PPDM achieves the best overall performance, with the lowest NMSE of 0.0223, the highest SSIM of 0.9880, and the highest PSNR of 30.868 dB, while requiring only 16.90 GB of memory. These results demonstrate that PPDM provides the best balance between reconstruction fidelity and computational efficiency for the PET denoising task.


\subsection{Ablation Studies}

\begin{table*}[htb!]
\tiny
\centering
\caption{Ablation study on spatial down-scale factor $r$ on Yale PET head denoising and attenuation correction task. We compare 3D BBDM and the proposed PPDM with different spatial down-scale factor $r$ using NMSE, SSIM, PSNR, training GPU memory consumption, and inference TFLOPs. Best results are highlighted in \textbf{bold}.}

\label{tab:eval_ldpet_pethead_models}
\resizebox{0.88\textwidth}{!}{
\begin{tabular}{l|c|c|c|c|c}
    \hline
    Model & NMSE & SSIM & PSNR & Training Memory(GB) & TFLOPs\Tstrut\Bstrut\\
    \hline
    3D BBDM              & $0.0131 \pm  0.0043$ & $0.9852 \pm 0.0047$ & $32.774 \pm 2.501$& $49.07$ &$18160$ \\
    PPDM($r=2$)      & $\mathbf{0.0130 \pm 0.0040}$ & $\mathbf{0.9854 \pm 0.0043}$ & $\mathbf{32.774 \pm 2.515}$& $6.94$ &$2280$\Tstrut\\
    PPDM($r=3$)      & $0.0137 \pm 0.0042$ & $0.9845 \pm 0.0045$ & $32.540 \pm 2.575$& $2.81$ &$651$\Tstrut\\
    \hline
\end{tabular}
}
\end{table*}

\textbf{Impact of spatial down-scale factor} $r$ in PPDM is summarized in Table~\ref{tab:eval_ldpet_pethead_models}. We evaluate PPDM with spatial down-scale factors $r = 1, 2, 3$, where $r=1$ corresponds to the original 3D BBDM without pixel puzzle operations. The performance of PPDM with $r=1$ and $r=2$ is comparable, with $r=2$ achieving a slight improvement in NMSE from 0.0131 to 0.0130. Meanwhile, increasing $r$ leads to a substantial reduction in computational cost: the training GPU memory decreases from 49.07 GB to 6.94 GB (approximately a $7\times$ reduction), and the inference TFLOPs drop from 18{,}160 to 2{,}280. When further increasing the down-scale factor to $r=3$, performance degrades moderately, with NMSE and PSNR changing from 0.0130 and 32.774 dB to 0.0137 and 32.540 dB, respectively. Nevertheless, PPDM with $r=3$ still outperforms other state-of-the-art methods reported in Table~\ref{tab:compare_method}, while achieving further reductions in training memory (2.81\,GB) and inference TFLOPs (651).

\begin{table}[htb!]
\footnotesize
\centering
\caption{Ablation study on puzzle-gradient kernel size on the IXI dataset.
Quantitative results for different Sobel kernel sizes used in the puzzle-gradient loss, evaluated using NMSE, SSIM, and PSNR. Best results are marked in \textbf{bold}.}
\label{tab:abl_IXI_kernalsize}
\begin{tabular}{l|c|c|c}
    \hline
    n = & NMSE & SSIM & PSNR \Tstrut\Bstrut\\
    \hline
    3      & $\mathbf{0.0059 \pm 0.0013}$ & $\mathbf{0.9934 \pm 0.0010}$ & $\mathbf{32.436 \pm 1.899}$\\
    5      & $0.0089 \pm 0.0016$ & $0.9902 \pm 0.0015$ & $30.676 \pm 2.023$ \Tstrut\\
    7      & $0.0091 \pm 0.0016$ & $0.9907 \pm 0.0013$ & $30.575 \pm 2.019$\\
    \hline
\end{tabular}
\end{table}

\begin{figure}[htb!]
\centering
\includegraphics[width=0.48\textwidth]{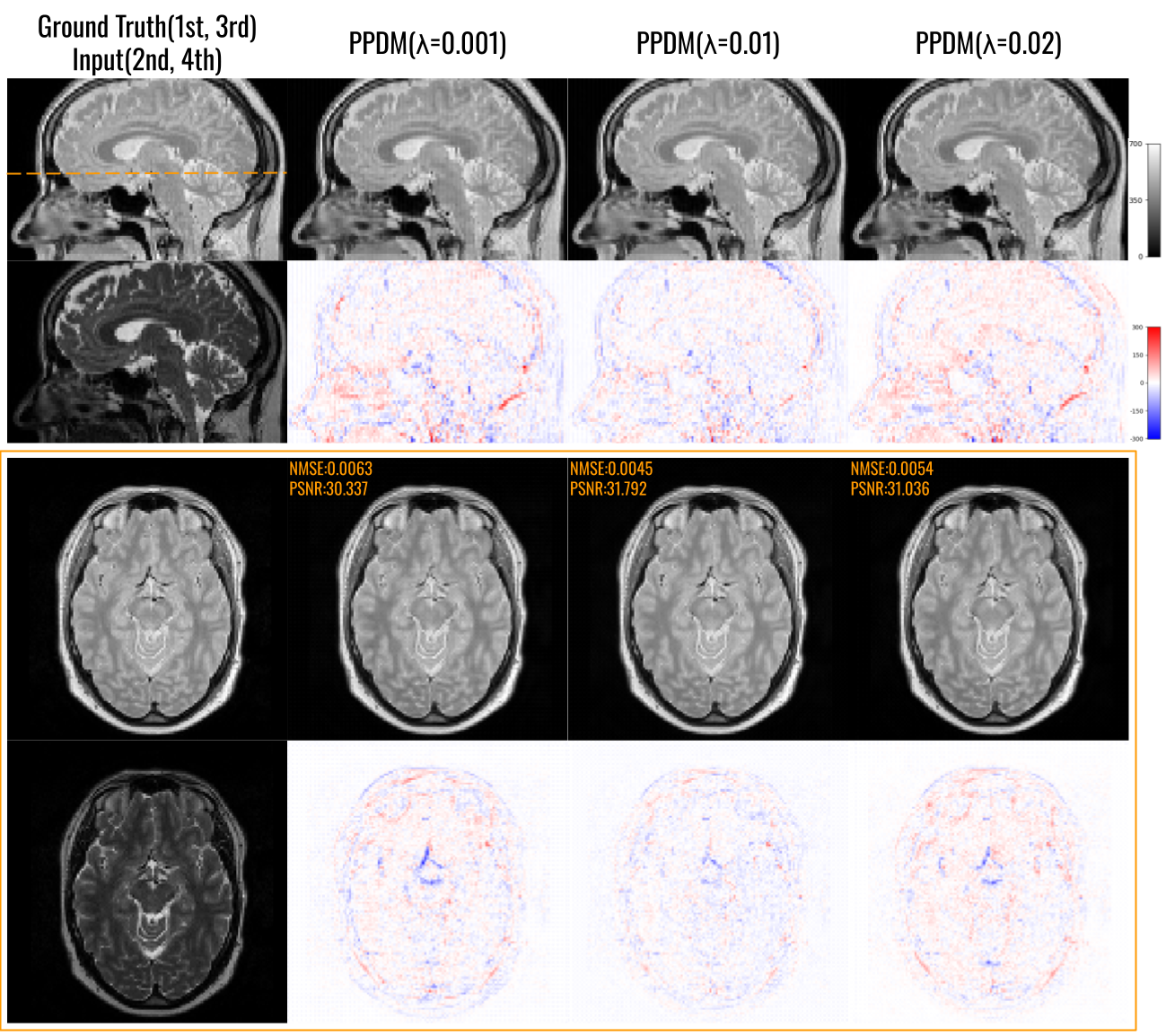}
\caption{Visual ablation study of PD-weighted MRI generated from T2-weighted MRI using different puzzle-gradient loss weights $\lambda$. Coronal slices (first row) with corresponding error maps (second row), and axial slices (third row) with corresponding error maps (fourth row), are shown. NMSE and PSNR are computed over the 3D volumes and reported at the top of the axial slices.}
\label{fig:abl_figure_lam}
\end{figure}

\textbf{Impact of the puzzle gradient loss factor $\lambda$} in PPDM was studied and is summarized in Fig.~\ref{fig:abl_plot_lam}. We investigated PPDM's performance with different values of $\lambda$ applied in the training loss. As $\lambda$ increases from 0.001 to 0.01, the NMSE first decreases from 0.0083 to 0.0056. However, as $\lambda$ continues to increase, the NMSE begins to rise again, reaching 0.0079 when $\lambda = 0.05$. Figure~\ref{fig:abl_figure_lam} shows the visual comparison of results generated from models trained with $\lambda = 0.001$, $0.01$, and $0.02$. When $\lambda = 0.001$, there is a significant grid-like artifact in both the reconstructed image and the error map, and the NMSE remains relatively high at 0.0063. When $\lambda$ increases to 0.01, the grid-like artifact becomes visually negligible. The NMSE decreases to 0.0045. As $\lambda$ continues to increase to 0.02, the grid artifact is removed from the background of the error map, but the accuracy slightly decreases, with an NMSE of 0.0054. 

\textbf{Impact of the kernel size} in PPDM was studied and is summarized in Table~\ref{tab:abl_IXI_kernalsize}. For the IXI dataset with a puzzle factor of $r = 2$, increasing the kernel size of the puzzle gradient $E(X)$ from 3 to 5 and 7 leads to a consistent degradation in performance: the NMSE rises from 0.0059 to 0.0089 and 0.0091, respectively, indicating that larger kernels introduce excessive smoothing and reduce reconstruction fidelity. However, when the puzzle factor increases to $r = 3$, as shown in Table~\ref{tab:abl_PET_head_lam3_kernalsize}, the trend changes. In this setting, the best performance is achieved at kernel size = 5, yielding the lowest NMSE (0.0137) and the highest SSIM and PSNR. These observations suggest that the optimal kernel size depends on the puzzle factor: smaller kernels are preferable when $r = 2$, whereas moderate kernel sizes offer better accuracy when $r = 3$.

\begin{table}[htb!]
\footnotesize
\centering
\caption{ablative study of puzzle-gradient kernel size under higher spatial down-scale factor ($r=3$) on Yale PET head 3D (5\%). Best results are marked in \textbf{bold}.}
\label{tab:abl_PET_head_lam3_kernalsize}
\begin{tabular}{l|c|c|c}
    \hline
    n =  & NMSE & SSIM & PSNR \Tstrut\Bstrut\\
    \hline
    3      & $0.0152 \pm 0.0043$ & $0.9824 \pm 0.0049$ & $32.051 \pm 2.446$ \Tstrut\\
    5      & $\mathbf{0.0137 \pm 0.0042}$ & $\mathbf{0.9845 \pm 0.0045}$ & $\mathbf{32.540 \pm 2.575}$\\
    7      & $0.0145 \pm 0.0041$ & $0.9839 \pm 0.0045$ & $32.255 \pm 2.4813$\\
    \hline
\end{tabular}
\end{table}

\textbf{Impact of kernel direction} in PPDM is summarized in Table~\ref{tab:kernal_dir}. We evaluate the performance of PPDM when trained with puzzle-gradient losses constrained by Sobel kernels applied along different spatial directions. When only a single kernel direction is used, the NMSE remains in the range of 0.0077--0.0081, indicating limited performance improvement. Incorporating two directional kernels yields nearly identical results, suggesting that partial directional constraints are insufficient to enforce full volumetric structural consistency. In contrast, constraining kernels along all three spatial directions achieves the best performance, yielding the lowest NMSE of 0.0059.

\begin{table} [htb!]
\tiny
\centering
\caption{Ablative studies on the inclusion of multi-directional kernel in PPDM. \textcolor{green}{\cmark} and \xmark\space means used and not used the kernel of direction in PPDM training, respectively. Best results are marked in \textbf{bold}.}
\label{tab:kernal_dir}
\resizebox{0.45 \textwidth}{!}{
\begin{tabular}{|c c c ||c|c|c|}
    \hline
    x                   & y                & z               & PSNR                   & NMSE               \Tstrut\Bstrut\\
    \hline   
    \textcolor{green}{\cmark}    & \xmark                      & \xmark                      & $31.314 \pm 1.906$         & $0.0077 \pm 0.0017$    \Tstrut\Bstrut\\
    \xmark                       & \textcolor{green}{\cmark}   & \xmark                      & $31.101 \pm 2.002$         & $0.0080  \pm 0.0015$    \Tstrut\Bstrut\\
    \xmark                       & \xmark                      & \textcolor{green}{\cmark}   & $31.331 \pm 1.896$         & $0.0077 \pm 0.0018$    \Tstrut\Bstrut\\
    \textcolor{green}{\cmark}    & \textcolor{green}{\cmark}   & \xmark                      & $31.113 \pm 1.895$         & $0.0081 \pm 0.0019$    \Tstrut\Bstrut\\
    \xmark                       & \textcolor{green}{\cmark}   & \textcolor{green}{\cmark}   & $31.313 \pm 1.927$         & $0.0077 \pm 0.0016$    \Tstrut\Bstrut\\
    \textcolor{green}{\cmark}    & \xmark                      & \textcolor{green}{\cmark}   & $31.103\pm 1.929$         & $0.0081 \pm 0.0017$    \Tstrut\Bstrut\\
    \textcolor{green}{\cmark}    & \textcolor{green}{\cmark}   & \textcolor{green}{\cmark}   & $\mathbf{32.436 \pm 1.899}$   & $\mathbf{0.0059 \pm 0.0013}$    \Tstrut\Bstrut\\
    \hline
\end{tabular}
}
\end{table}

\section{Discussion} 
In this work, we introduce a memory-efficient diffusion framework, the Pixel Puzzle Diffusion Model (PPDM), which substantially reduces GPU memory consumption for 3D diffusion-based medical image translation while achieving equal or superior performance compared to the 3D BBDM. During training, the input volume is spatially unpuzzled into multiple lower-resolution channels, allowing the diffusion network to operate on reduced spatial dimensions without losing global volumetric context. The model directly learns the mapping from the input to the target within $T$ denoising steps, and a puzzle gradient loss is incorporated to enhance edge preservation and anatomical fidelity. During inference, the input is similarly unpuzzled, processed through the diffusion steps, and subsequently puzzled back to reconstruct the full-resolution output volume. There are several key advantages of the proposed PPDM framework. First, 3D diffusion-based image-to-image translation typically demands substantial GPU memory, especially for high-resolution volumes. PPDM employs a reversible pixel puzzle-unpuzzle operator to reduce spatial resolution while preserving global volumetric context, enabling efficient training on large 3D medical imaging tasks using limited computational resources. As shown in Table~\ref{tab:compare_method}, the minimum GPU memory required for training decreases from nearly 50 GB with a 3D BBDM to only 6.94 GB with PPDM. Although methods, such as WDM and ALDM, also operate under lower memory budgets, they exhibit considerable performance degradation compared with both PPDM and the 3D BBDM baseline, while requiring a reliable encoder/decoder for volume compression to be trained first. Second, PPDM introduces a puzzle gradient loss during training, which enhances structural fidelity and reduces the grid-like artifact caused by channel mismatch, thereby improving fine anatomical detail recovery. Third, PPDM is inherently scalable. By adjusting the puzzle factor $r$, spatial resolution can be flexibly reduced by a factor of 2, 3, or higher, enabling controllable trade-offs between memory usage and accuracy. As demonstrated in Table~\ref{tab:eval_ldpet_pethead_models}, even with increased compression, PPDM maintains competitive performance, demonstrating strong robustness across different compression ratios. Last and most importantly, PPDM remains simple and efficient by strategically integrating the puzzle strategy with gradient loss constraints.

Our method is broadly applicable to a wide range of 3D medical image–to–image translation tasks. By substantially reducing GPU memory requirements during both training and inference, it enables the use of higher-resolution volumetric data and makes large-scale 3D modeling more accessible. As demonstrated in our experiments, the proposed framework can be effectively applied to PET denoising, joint PET denoising and attenuation correction, and MRI sequence translation. Across all tasks, our method consistently achieves the best performance among state-of-the-art approaches, highlighting its robustness, efficiency, and generalizability in diverse 3D reconstruction and translation settings.

However, our PPDM method still has several limitations that warrant further investigation. First, the model can be relatively sensitive to training hyperparameters. As shown in Fig.~\ref{fig:abl_plot_lam}, small changes in the loss weight $\lambda$ can lead to noticeable variations in performance, and the optimal value of $\lambda$ may differ slightly across applications. Nevertheless, despite this sensitivity, PPDM requires substantially fewer computational resources for both training and sampling, making it considerably faster to train than other state-of-the-art diffusion approaches. Second, as the spatial down-scale factor $r$ increases, the performance of PPDM also gradually decreases. This is a trade-off between performance and memory/speed saving. When $r$ is increased to 3, the minimum training memory decreases by more than 20× compared with the original diffusion model, while the NMSE only increases slightly from 0.0131 to 0.0137, still outperforming WDM, despite their larger model sizes. Moreover, as shown in Table~\ref{tab:abl_PET_head_lam3_kernalsize}, larger spatial down-scale factors require careful adjustment of the kernel size to achieve optimal performance, which adds complexity to the training process. This sensitivity may arise because small kernels (e.g., size 3) are insufficient to fully capture the puzzle–unpuzzle process, leading to weaker synchronization among channels, whereas larger kernels (e.g., size 7) oversmooth edge gradients and also fail to provide the best performance. Developing a more adaptive or unified approach that maintains strong performance across different spatial down-scale factors is an interesting direction for future work.


\section{Conclusion} 
We introduced the Pixel Puzzling Diffusion Model (PPDM), a memory- and speed-efficient framework for 3D diffusion-based medical image translation. By employing a reversible pixel puzzle–unpuzzle operator, PPDM significantly reduces GPU memory consumption while preserving global volumetric context. In combination with a direct bridge diffusion formulation and a puzzle-gradient loss that enforces spatial coherence, PPDM mitigates channel-wise inconsistencies and avoids grid-like artifacts during reconstruction. Extensive experiments on challenging PET and MRI translation tasks demonstrate that PPDM consistently matches or surpasses the performance of full 3D diffusion models while requiring substantially fewer computational resources. These results indicate that PPDM provides a practical and scalable solution for high-resolution volumetric diffusion modeling, facilitating broader deployment of diffusion-based techniques in medical imaging applications.

\section*{Acknowledgments}
Parts of the data used in preparation of this article were obtained from the University of Bern, Dept. of Nuclear Medicine and School of Medicine, Ruijin Hospital. As such, the investigators contributed to the design and implementation of DATA and/or provided data but did not participate in analysis or writing of this report. A complete listing of investigators can be found at: “https://udpet-challenge.github.io/”

\section*{CRediT authorship contribution statement}
\textbf{Tianqi Chen}: Writing – original draft, Visualization, Validation, Software, Project administration, Methodology, Investigation, Conceptualization. 
\textbf{Jun Hou}: Formal analysis, Writing – review \& editing. 
\textbf{Yinchi Zhou}: Writing – review \& editing. 
\textbf{James S. Duncan}: Writing – review \& editing. 
\textbf{Chi Liu}: Writing – review \& editing. 
\textbf{Bo Zhou}: Writing – review \& editing, Supervision, Project administration, Methodology, Investigation, Conceptualization.

\section*{Declaration of competing interest}
The authors declare that they have no known competing financial interests or personal relationships that could have appeared to influence the work reported in this paper.

\bibliographystyle{model2-names.bst}\biboptions{authoryear}
\bibliography{refs}

@inproceedings{kim2024adaptive,
  title={Adaptive latent diffusion model for 3d medical image to image translation: Multi-modal magnetic resonance imaging study},
  author={Kim, Jonghun and Park, Hyunjin},
  booktitle={Proceedings of the IEEE/CVF Winter conference on applications of computer Vision},
  pages={7604--7613},
  year={2024}
}

@article{huang2024wavedm,
  title={Wavedm: Wavelet-based diffusion models for image restoration},
  author={Huang, Yi and Huang, Jiancheng and Liu, Jianzhuang and Yan, Mingfu and Dong, Yu and Lv, Jiaxi and Chen, Chaoqi and Chen, Shifeng},
  journal={IEEE Transactions on Multimedia},
  volume={26},
  pages={7058--7073},
  year={2024},
  publisher={IEEE}
}

@article{ho2020denoising,
  title={Denoising diffusion probabilistic models},
  author={Ho, Jonathan and Jain, Ajay and Abbeel, Pieter},
  journal={Advances in neural information processing systems},
  volume={33},
  pages={6840--6851},
  year={2020}
}

@inproceedings{isola2017image,
  title={Image-to-image translation with conditional adversarial networks},
  author={Isola, Phillip and Zhu, Jun-Yan and Zhou, Tinghui and Efros, Alexei A},
  booktitle={Proceedings of the IEEE conference on computer vision and pattern recognition},
  pages={1125--1134},
  year={2017}
}

@article{gong2020parameter,
  title={Parameter-transferred Wasserstein generative adversarial network (PT-WGAN) for low-dose PET image denoising},
  author={Gong, Yu and Shan, Hongming and Teng, Yueyang and Tu, Ning and Li, Ming and Liang, Guodong and Wang, Ge and Wang, Shanshan},
  journal={IEEE transactions on radiation and plasma medical sciences},
  volume={5},
  number={2},
  pages={213--223},
  year={2020},
  publisher={IEEE}
}

@inproceedings{hu2020simultaneous,
  title={Simultaneous attenuation correction, scatter correction, and denoising in pet imaging with deep learning},
  author={Hu, Jicun and Whiteley, William and Zhang, Xiang and Zhou, Chuanyu and Panin, Vladimir},
  booktitle={2020 IEEE Nuclear Science Symposium and Medical Imaging Conference (NSS/MIC)},
  pages={1--3},
  year={2020},
  organization={IEEE}
}

@article{nie2018medical,
  title={Medical image synthesis with deep convolutional adversarial networks},
  author={Nie, Dong and Trullo, Roger and Lian, Jun and Wang, Li and Petitjean, Caroline and Ruan, Su and Wang, Qian and Shen, Dinggang},
  journal={IEEE Transactions on Biomedical Engineering},
  volume={65},
  number={12},
  pages={2720--2730},
  year={2018},
  publisher={IEEE}
}

@article{yang2020mri,
  title={MRI cross-modality image-to-image translation},
  author={Yang, Qianye and Li, Nannan and Zhao, Zixu and Fan, Xingyu and Chang, Eric I-Chao and Xu, Yan},
  journal={Scientific reports},
  volume={10},
  number={1},
  pages={3753},
  year={2020},
  publisher={Nature Publishing Group UK London}
}

@article{li2022tcgan,
  title={TCGAN: a transformer-enhanced GAN for PET synthetic CT},
  author={Li, Jitao and Qu, Zongjin and Yang, Yue and Zhang, Fuchun and Li, Meng and Hu, Shunbo},
  journal={Biomedical Optics Express},
  volume={13},
  number={11},
  pages={6003--6018},
  year={2022},
  publisher={Optica Publishing Group}
}

@article{bazangani2022fdg,
  title={FDG-PET to T1 weighted MRI translation with 3D elicit generative adversarial network (E-GAN)},
  author={Bazangani, Farideh and Richard, Fr{\'e}d{\'e}ric JP and Ghattas, Badih and Guedj, Eric},
  journal={Sensors},
  volume={22},
  number={12},
  pages={4640},
  year={2022},
  publisher={MDPI}
}

@article{gong2024pet,
  title={PET image denoising based on denoising diffusion probabilistic model},
  author={Gong, Kuang and Johnson, Keith and El Fakhri, Georges and Li, Quanzheng and Pan, Tinsu},
  journal={European Journal of Nuclear Medicine and Molecular Imaging},
  volume={51},
  number={2},
  pages={358--368},
  year={2024},
  publisher={Springer}
}

@article{chen20252,
  title={2.5 D multi-view averaging diffusion model for 3D medical image translation: application to low-count PET reconstruction with CT-less attenuation correction},
  author={Chen, Tianqi and Hou, Jun and Zhou, Yinchi and Xie, Huidong and Chen, Xiongchao and Liu, Qiong and Guo, Xueqi and Xia, Menghua and Duncan, James S and Liu, Chi and others},
  journal={IEEE Transactions on Medical Imaging},
  year={2025},
  publisher={IEEE}
}

@inproceedings{xing2024cross,
  title={Cross-conditioned diffusion model for medical image to image translation},
  author={Xing, Zhaohu and Yang, Sicheng and Chen, Sixiang and Ye, Tian and Yang, Yijun and Qin, Jing and Zhu, Lei},
  booktitle={International Conference on Medical Image Computing and Computer-Assisted Intervention},
  pages={201--211},
  year={2024},
  organization={Springer}
}

@inproceedings{ronneberger2015u,
  title={U-net: Convolutional networks for biomedical image segmentation},
  author={Ronneberger, Olaf and Fischer, Philipp and Brox, Thomas},
  booktitle={International Conference on Medical image computing and computer-assisted intervention},
  pages={234--241},
  year={2015},
  organization={Springer}
}

@inproceedings{li2023bbdm,
  title={Bbdm: Image-to-image translation with brownian bridge diffusion models},
  author={Li, Bo and Xue, Kaitao and Liu, Bin and Lai, Yu-Kun},
  booktitle={Proceedings of the IEEE/CVF conference on computer vision and pattern Recognition},
  pages={1952--1961},
  year={2023}
}

@article{zhou2021limited,
  title={Limited view tomographic reconstruction using a cascaded residual dense spatial-channel attention network with projection data fidelity layer},
  author={Zhou, Bo and Zhou, S Kevin and Duncan, James S and Liu, Chi},
  journal={IEEE transactions on medical imaging},
  volume={40},
  number={7},
  pages={1792--1804},
  year={2021},
  publisher={IEEE}
}

@article{zhang2018sparse,
  title={A sparse-view CT reconstruction method based on combination of DenseNet and deconvolution},
  author={Zhang, Zhicheng and Liang, Xiaokun and Dong, Xu and Xie, Yaoqin and Cao, Guohua},
  journal={IEEE transactions on medical imaging},
  volume={37},
  number={6},
  pages={1407--1417},
  year={2018},
  publisher={IEEE}
}

@article{welander2018generative,
  title={Generative adversarial networks for image-to-image translation on multi-contrast mr images-a comparison of cyclegan and unit},
  author={Welander, Per and Karlsson, Simon and Eklund, Anders},
  journal={arXiv preprint arXiv:1806.07777},
  year={2018}
}

@inproceedings{zhou2020dudornet,
  title={DuDoRNet: learning a dual-domain recurrent network for fast MRI reconstruction with deep T1 prior},
  author={Zhou, Bo and Zhou, S Kevin},
  booktitle={Proceedings of the IEEE/CVF conference on computer vision and pattern recognition},
  pages={4273--4282},
  year={2020}
}

@article{zhou2024pour,
  title={POUR-Net: A Population-Prior-Aided Over-Under-Representation Network for Low-Count PET Attenuation Map Generation},
  author={Zhou, Bo and Hou, Jun and Chen, Tianqi and Zhou, Yinchi and Chen, Xiongchao and Xie, Huidong and Liu, Qiong and Guo, Xueqi and Tsai, Yu-Jung and Panin, Vladimir Y and others},
  journal={arXiv preprint arXiv:2401.14285},
  year={2024}
}

@article{xie2024dose,
  title={Dose-aware Diffusion Model for 3D PET Image Denoising: Multi-institutional Validation with Reader Study and Real Low-dose Data},
  author={Xie, Huidong and Gan, Weijie and Bayerlein, Reimund and Zhou, Bo and Chen, Ming-Kai and Kulon, Michal and Boustani, Annemarie and Ko, Kuan-Yin and Wang, Der-Shiun and Spencer, Benjamin A and others},
  journal={arXiv preprint arXiv:2405.12996},
  year={2024}
}

@inproceedings{wolterink2017deep,
  title={Deep MR to CT synthesis using unpaired data},
  author={Wolterink, Jelmer M and Dinkla, Anna M and Savenije, Mark HF and Seevinck, Peter R and van den Berg, Cornelis AT and I{\v{s}}gum, Ivana},
  booktitle={International workshop on simulation and synthesis in medical imaging},
  pages={14--23},
  year={2017},
  organization={Springer}
}

@article{shafai2019dose,
  title={Dose evaluation of MRI-based synthetic CT generated using a machine learning method for prostate cancer radiotherapy},
  author={Shafai-Erfani, Ghazal and Wang, Tonghe and Lei, Yang and Tian, Sibo and Patel, Pretesh and Jani, Ashesh B and Curran, Walter J and Liu, Tian and Yang, Xiaofeng},
  journal={Medical Dosimetry},
  volume={44},
  number={4},
  pages={e64--e70},
  year={2019},
  publisher={Elsevier}
}

@article{palmer2021synthetic,
  title={Synthetic computed tomography data allows for accurate absorbed dose calculations in a magnetic resonance imaging only workflow for head and neck radiotherapy},
  author={Palm{\'e}r, Emilia and Karlsson, Anna and Nordstr{\"o}m, Fredrik and Petruson, Karin and Siversson, Carl and Ljungberg, Maria and Sohlin, Maja},
  journal={Physics and imaging in radiation oncology},
  volume={17},
  pages={36--42},
  year={2021},
  publisher={Elsevier}
}

@article{aouadi2025towards,
  title={Towards MR-Only radiotherapy in head and neck: generation of synthetic CT from Zero-TE MRI using deep learning},
  author={Aouadi, Souha and Barzegar, Mojtaba and Al-Sabahi, Alla and Torfeh, Tarraf and Paloor, Satheesh and Riyas, Mohamed and Caparrotti, Palmira and Hammoud, Rabih and Al-Hammadi, Noora},
  journal={Information},
  volume={16},
  number={6},
  pages={477},
  year={2025},
  publisher={MDPI}
}

@article{vellini2025deep,
  title={A deep learning algorithm to generate synthetic computed tomography images for brain treatments from 0.35 T magnetic resonance imaging},
  author={Vellini, Luca and Quaranta, Flaviovincenzo and Menna, Sebastiano and Pilloni, Elisa and Catucci, Francesco and Lenkowicz, Jacopo and Votta, Claudio and Aquilano, Michele and D’Aviero, Andrea and Iezzi, Martina and others},
  journal={Physics and Imaging in Radiation Oncology},
  volume={33},
  pages={100708},
  year={2025},
  publisher={Elsevier}
}

@article{lyu2022conversion,
  title={Conversion between ct and mri images using diffusion and score-matching models},
  author={Lyu, Qing and Wang, Ge},
  journal={arXiv preprint arXiv:2209.12104},
  year={2022}
}

@article{liu20232,
  title={I$^{2}$\uppercase{SB}: Image-to-Image Schr\"{o}dinger Bridge},
  author={Liu, Guan-Horng and Vahdat, Arash and Huang, De-An and Theodorou, Evangelos A and Nie, Weili and Anandkumar, Anima},
  journal={arXiv preprint arXiv:2302.05872},
  year={2023}
}

@inproceedings{shen2023pet,
  title={PET image denoising with score-based diffusion probabilistic models},
  author={Shen, Chenyu and Yang, Ziyuan and Zhang, Yi},
  booktitle={International Conference on Medical Image Computing and Computer-Assisted Intervention},
  pages={270--278},
  year={2023},
  organization={Springer}
}

@article{pan2024full,
  title={Full-dose whole-body PET synthesis from low-dose PET using high-efficiency denoising diffusion probabilistic model: PET consistency model},
  author={Pan, Shaoyan and Abouei, Elham and Peng, Junbo and Qian, Joshua and Wynne, Jacob F and Wang, Tonghe and Chang, Chih-Wei and Roper, Justin and Nye, Jonathon A and Mao, Hui and others},
  journal={Medical Physics},
  volume={51},
  number={8},
  pages={5468--5478},
  year={2024},
  publisher={Wiley Online Library}
}

@article{graf2023denoising,
  title={Denoising diffusion-based MRI to CT image translation enables automated spinal segmentation},
  author={Graf, Robert and Schmitt, Joachim and Schlaeger, Sarah and M{\"o}ller, Hendrik Kristian and Sideri-Lampretsa, Vasiliki and Sekuboyina, Anjany and Krieg, Sandro Manuel and Wiestler, Benedikt and Menze, Bjoern and Rueckert, Daniel and others},
  journal={European Radiology Experimental},
  volume={7},
  number={1},
  pages={70},
  year={2023},
  publisher={Springer}
}

@inproceedings{choo2024slice,
  title={Slice-consistent 3d volumetric brain ct-to-mri translation with 2d brownian bridge diffusion model},
  author={Choo, Kyobin and Jun, Youngjun and Yun, Mijin and Hwang, Seong Jae},
  booktitle={International Conference on Medical Image Computing and Computer-Assisted Intervention},
  pages={657--667},
  year={2024},
  organization={Springer}
}

@misc{IXIDataset2018,
  author       = {{Biomedical Image Analysis Group} and {Imperial College London} and {Centre for the Developing Brain} and {King's College London}},
  title        = {Information eXtraction From Images (IXI) Dataset},
  year         = {2018},
  howpublished = {\url{https://brain-development.org/ixi-dataset/}},
}

@article{xue2022cross,
  title={A cross-scanner and cross-tracer deep learning method for the recovery of standard-dose imaging quality from low-dose PET},
  author={Xue, Song and Guo, Rui and Bohn, Karl Peter and Matzke, Jared and Viscione, Marco and Alberts, Ian and Meng, Hongping and Sun, Chenwei and Zhang, Miao and Zhang, Min and others},
  journal={European journal of nuclear medicine and molecular imaging},
  volume={49},
  number={6},
  pages={1843--1856},
  year={2022},
  publisher={Springer}
}

@article{xue2025deep,
  title={A deep learning method for the recovery of standard-dose imaging quality from ultra-low-dose PET on wavelet domain},
  author={Xue, Song and Liu, Fanxuan and Wang, Hanzhong and Zhu, Hong and Sari, Hasan and Viscione, Marco and Sznitman, Raphael and Rominger, Axel and Guo, Rui and Li, Biao and others},
  journal={European Journal of Nuclear Medicine and Molecular Imaging},
  volume={52},
  number={5},
  pages={1901--1911},
  year={2025},
  publisher={Springer}
}

@inproceedings{bieder2024memory,
  title={Memory-efficient 3d denoising diffusion models for medical image processing},
  author={Bieder, Florentin and Wolleb, Julia and Durrer, Alicia and Sandkuehler, Robin and Cattin, Philippe C},
  booktitle={Medical Imaging with Deep Learning},
  pages={552--567},
  year={2024},
  organization={PMLR}
}

\end{document}